\pdfoutput=1
\documentclass[11pt]{article}

\usepackage{ACL2023}

\usepackage{times}
\usepackage{latexsym}

\usepackage[T1]{fontenc}
\usepackage[utf8]{inputenc}

\usepackage{microtype}

\usepackage{inconsolata}
\usepackage{graphics}
\usepackage{graphicx}
\usepackage{multirow}
\usepackage[vlined]{algorithm2e}
\usepackage{amsfonts}

\title{Matching Tasks with Industry Groups for Augmenting Commonsense Knowledge}

\author{Rituraj Singh\thanks{~~Work done while working at TCS Research} \and Sachin Pawar \and Girish K. Palshikar \\
  TCS Research, Tata Consultancy Services Limited, India. \\
  \texttt{riturajsingh.infotech@gmail.com, sachin7.p@tcs.com, gk.palshikar@tcs.com} \\}

\begin{document}
\maketitle
\begin{abstract}
Commonsense knowledge bases (KB) are a source of specialized knowledge %and auxiliary information 
that is widely used to improve machine learning applications. However, even for a large KB such as ConceptNet, capturing explicit knowledge from each industry domain is challenging. For example, only a few samples of general {\em tasks} performed by various industries are available in ConceptNet. Here, a task is a well-defined knowledge-based volitional action to achieve a particular goal. In this paper, we aim to fill this gap and present a weakly-supervised framework to augment commonsense KB with tasks carried out by various industry groups (IG). 
%First, we automatically extract tasks from the text using a BERT-based weakly supervised neural model. Next, we apply an unsupervised neural approach to label each task to an IG. %Transforming each task to its canonical form, 
%We then train a neural model to learn task-IG affinity and apply clustering to select the top-k tasks per IG. 
We attempt to {\em match} each task with one or more suitable IGs by training
%We train 
a neural model to learn task-IG affinity and apply clustering to select the top-k tasks per IG. 
%We extract a total of 2339 triples of the form $\langle IG, is~capable~of, task \rangle$ from two publicly available news datasets for 24 IGs. The proposed methodology yields a precision of 0.86 that validates the reliability of the extracted task-IG pairs that can be directly added to existing KBs.
We extract a total of 2339 triples of the form $\langle IG, is~capable~of, task \rangle$ from two publicly available news datasets for 24 IGs with the precision of 0.86. This validates the reliability of the extracted task-IG pairs that can be directly added to existing KBs.
\end{abstract}

\section{Introduction}
Several NLP applications take advantage of common sense knowledge such as question answering~\cite{zhong2019improving,feng2020scalable}, textual entailment~\cite{kapanipathi2020infusing}, sentiment analysis~\cite{agarwal2015sentiment}, and summarization~\cite{gerani2019modeling}. In this paper, we focus on a specific type of knowledge -- {\em tasks} performed by organizations belonging to a certain {\em industry group} (IG). Here, a task is a well-defined knowledge-based action that is carried out volitionally~\cite{pawar2021weakly}. 
%Determining the industry to which a task belongs has intermediate and numerous real-life end-user applications. NLP-based applications may be improved by providing it with specialized knowledge and external information sources. 
We advocate augmenting an existing common-sense knowledge base (KB) with the knowledge of several tasks being performed by an industry group. For example, ConceptNet~\cite{speer2017conceptnet} consists of common sense knowledge about the world in the form of triples such as $\langle${\small\texttt{Doctor}}, \textit{is~capable~of}, {\small\tt help~a~sick~person}$\rangle$. However, it is observed that ConceptNet has very limited common sense knowledge about industry groups. For instance, %let us consider an example of an industry group {\small\texttt{Telecommunication}}. The 
the ConceptNet node for {\small\sf Power company}\footnote{https://conceptnet.io/c/en/power\_company} 
consists of no knowledge of the tasks being carried out in the energy industry. Often, such knowledge is limited or largely absent for most industry groups (see Section~\ref{secCN}). In this work, we propose to overcome such limitations by augmenting considerable knowledge to common-sense KBs %. For example, we advocate to augment knowledge 
of the form $\langle${\small\sf Energy company}, \textit{is~capable~of}, {\small\tt operate~coal~fired~plants}$\rangle$. %Such augments will improve the general knowledge about tasks performed by different industry groups, 
It will help NLP applications by allowing them to better understand the working of industry groups.%~\citep{kapanipathi2020infusing,zhong2019improving,agarwal2015sentiment}.

%Effectively organising and using common sense knowledge to understand the working of the various industries is a challenge. 
In this work, we use the GICS\footnote{https://www.msci.com/our-solutions/indexes/gics} standard taxonomy that classifies organizations into 24 \emph{industry groups} (IGs) %considering produced goods, manufacturing processes, %financial market trends 
%and the services provided. %Each IG is characteristically different from the other. 
%Some example IGs are {\small\sf Banks}, {\small\sf Retailing}, {\small\sf Software \& Services}, etc. 
such as {\small\sf Banks}, {\small\sf Retailing}, {\small\sf Software \& Services}, etc.
%There are such 24 different IGs, refer appendix for the full list. %Perhaps the most important question for common sense knowledge for an IG is: what are the different tasks carried out by or within a particular industry group? A task is a well-defined knowledge-based action with a specific goal that is often voluntarily carried out by a single person, a group of persons, an organisation, a device or a system within a small time~\cite{pawar2021weakly}.
%The task extraction problem is closely related to event extraction problem which is a popular task in NLP. Event extraction is generally defined as specific occurrence of something happening at a certain time and place which involves one or more participants and often described as change of state. Although events are similar to task in few aspects, there are crucial differences such as tense formation, genericity, modularity, etc~\cite{pawar2021weakly}.Note that, to build the generic knowledge base, we focus on the \textit{tasks} carried out by industry groups which is inherently different from events.
%Information about tasks carried out by various IG resides in text documents. 
We propose to extract information about tasks carried out by various IGs from a large text corpus. For example, consider the news published by various business or general newspapers. A section of a news item may allude to a specific task that will be completed in future or is now being carried out by a certain organization (refer Table~\ref{tab: exampletask}). %Additionally, the organisation advertises their ongoing tasks in news for publicity and to attract new clients. Some examples of tasks include, {\small\tt{analyse bank transactions, install solar plus battery storage, create a realistic 3D avatar}}, etc. The classification of tasks based on different IG helps to organize, study and infer new knowledge. Hence, augmenting task-IG information is of utmost practical importance. In this work we consider the standard GICS\footnote{https://www.msci.com/our-solutions/indexes/gics} taxonomies to classify the tasks in corresponding 24 IG (For full list refer~\ref{tab: appendix_IG_hypothesis}).

%In general, an industry group performs a similar set of tasks to provide end services. For example, a {\small\texttt{Energy}} industry group does most tasks related to the installation of new energy sources, energy storage and exploration \& production of fuels, etc., while an {\small\texttt{Insurance}} group mostly carries tasks such as insurance broking, health insurance, casualty insurance, etc. 

\begin{table*}[t]
\scalebox{0.8}{
\begin{tabular}{|c|l|c|}
\hline
\textbf{Dataset} & \multicolumn{1}{c|}{\textbf{Tasks}}                                                                                                                                                                                                                                                                   & \textbf{Industry Group}                         \\ \hline
Reuters          & {\small\tt{Pacific Gas began \textbf{[construction of the two nuclear power units]}}} in 1969.                                                                                                                                                                                                                         & \multirow{2}{*}{Energy}                         \\ \cline{1-2}
TechCrunch        & \begin{tabular}[c]{@{}l@{}}{\small\tt{Total Petroleum NA ( TPN ) \textbf{{[}shut down several small crude oil pipelines{]}} operating near}}\\{\small\tt{the Texas/Oklahoma border last Friday as a precaution against damage from local flooding,}} \\ {\small\tt{according to Gary Zollinger , manager of operations.}}\end{tabular}       &                                                 \\ \hline
Reuters          & \begin{tabular}[c]{@{}l@{}}{\small\tt{Crews hoped to \textbf{{[}restore traffic to the line later today after clearing the damage train{]}}}}\\  {\small\tt{and repairing the tracks at Chacapalca , 225 km east of the Capital, Lima.}}\end{tabular}                                                                             & \multirow{2}{*}{Transportation}                 \\ \cline{1-2}
TechCrunch        & \begin{tabular}[c]{@{}l@{}}{\small\tt{We launched new tools for airlines so they can better \textbf{{[}predict consumer demand{]}} and \textbf{{[}plan}}} \\{\small\tt{\textbf{their routes{]}}}}.\end{tabular}                                                                                                                              &                                                 \\ \hline
%Reuters          & \begin{tabular}[c]{@{}l@{}}{\small\tt{Wolverine said it will \textbf{{[}concentrate its effort in the athletic footwear market in its }}}\\  {\small\tt{\textbf{Brooks footwear division{]}}.}}\end{tabular}                                                                                                                                       & \multirow{2}{*}{\begin{tabular}[c]{@{}c@{}}Consumer Durables \\and Apparrel\end{tabular}                                                                                                                                      } \\ \cline{1-2}
%TechCrunch        & \begin{tabular}[c]{@{}l@{}}{\small\tt{Intel, TSMC and Samsung Electronics are able to \textbf{{[}make chips of 10 - nanometers or lower{]}}}}\\ {\small\tt{,the fastest and most power - efficient chips currently on the market.}}\end{tabular}                                                                               &                                                 \\ \hline
Reuters          & \begin{tabular}[c]{@{}l@{}}{\small\tt{In other sectors , the Comet electrical chain \textbf{{[}raised retail profits{]}} by 46 pct to 17.4}}\\  {\small\tt{mln stg , while the Woolworth chain reported a 120 pct improvement to 38.7 mln.}}\end{tabular}                                                                               & \multirow{2}{*}{Retailing}                         \\ \cline{1-2}
TechCrunch        & \begin{tabular}[c]{@{}l@{}}{\small\tt{In those two months alone , Shopify seems to have \textbf{{[}onboarded more merchants{]}} than in the}} \\ {\small\tt{whole of 2018.}}\end{tabular}                                                                                                                                         &                                                 \\ \hline
Reuters          & \begin{tabular}[c]{@{}l@{}}{\small\tt{In November 1986 , Novo purchased 75 pct of shares in A/S Ferrosan,which heads a group}}\\ {\small\tt{specialising in \textbf{{[}research and development of CNS (central nervous treatment) treatments}}}\\  {\small\tt{\textbf{and the sale of pharmaceuticals and vitamins in Scandinavia{]}}}}.\end{tabular} & \multirow{2}{*}{Pharma}                         \\ \cline{1-2}
TechCrunch        & \begin{tabular}[c]{@{}l@{}}{\small\tt{Drugmaker Moderna has \textbf{{[}completed its initial efficacy analysis of its COVID - 19 vaccine{]}}}} \\{\small\tt{from the drug 's Phase 3 clinical study.}}\end{tabular}                                                                                                              &                                                 \\ \hline
Reuters          & \begin{tabular}[c]{@{}l@{}}{\small\tt{Laroche said he may \textbf{{[}obtain a short - term loan of up to one mln dlrs from Amoskeag Bank}}} \\ {\small\tt{\textbf{to help finance the purchase of shares under the offer{]}}}, bearing interest of up to nine pct.}\end{tabular}                                                              & \multirow{2}{*}{Banks}                          \\ \cline{1-2}
TechCrunch        & \begin{tabular}[c]{@{}l@{}}{\small\tt{Mobile banking startup Varo is becoming a real bank - The company announced that it has}} \\ {\small\tt{been \textbf{{[}granted a national bank charter from the Office of the Comptroller of the Currency{]}}}}.\end{tabular}                                                                     &                                                 \\ \hline
Reuters          & \begin{tabular}[c]{@{}l@{}}{\small\tt{The executives would also \textbf{{[}get cash settlements of options plans and continuation of}}} \\ {\small\tt{\textbf{insurance and other benefits{]}}.}}\end{tabular}                                                                                                                                    & \multirow{2}{*}{Insurance}                      \\ \cline{1-2}
TechCrunch        & \begin{tabular}[c]{@{}l@{}}{\small\tt{Square said it had also received approval from the FDIC to}} {\small\tt\textbf{{[}conduct deposit insurance{]}}}.\end{tabular}                                                                                                                                                     &                                                 \\ \hline
Reuters          & \begin{tabular}[c]{@{}l@{}}{\small\tt{Hogan said Systems  \textbf{{[}provides integrated applications software and processing services{]}}}}\\{\small\tt{to about 30 community banks}}.\end{tabular}                                                                                                                             & \multirow{2}{*}{\begin{tabular}[c]{@{}c@{}}Software and \\ Services\end{tabular}}          \\ \cline{1-2}
TechCrunch        & \begin{tabular}[c]{@{}l@{}}{\small\tt{Unlike its rivals , Zoox is \textbf{{[}developing the self - driving software stack, the on-demand}}} \\ {\small\tt{\textbf{ride - sharing app and the vehicle itself{]}}}}\end{tabular}                                                                                                                   &                                                 \\ \hline
Reuters          & \begin{tabular}[c]{@{}l@{}}{\small\tt{CIP , Canada 's second largest newsprint producer , recently \textbf{{[}launched a 366 mln Canadian }}} \\ {\small\tt{\textbf{dlr newsprint mill at Gold River , British Columbia{]}} which is due begin producing 230, 000}} \\ {\small\tt{metric tonnes per year by fall of 1989.}}\end{tabular}                       & \multirow{2}{*}{                                                                                                                       \begin{tabular}[c]{@{}c@{}} Media and \\Entertainment\end{tabular}}       \\ \cline{1-2}
TechCrunch        & \begin{tabular}[c]{@{}l@{}}{\small\tt{The company has also worked to \textbf{{[}create virtual creatures and characters in movies like}}}\\  {\small\tt{\textbf{ "The Lord of the Rings"{]}}.}}\end{tabular}                                                                                                                                       &                                                 \\ \hline
%Reuters          & \begin{tabular}[c]{@{}l@{}}{\small\tt{The Department said these falls were only partly offset by a rise in prices of home}}\\  {\small\tt{ - \textbf{{[}produced food manufacturing materials{]}}.}}\end{tabular}                                                                                                                         & \multirow{2}{*}{\begin{tabular}[c]{@{}c@{}}Food, Beverage\\ \& Tobacco\end{tabular}                                                                                                                        }      \\ \cline{1-2}
%TechCrunch        & \begin{tabular}[c]{@{}l@{}}{\small\tt{This will further aid boosting farmers income and \textbf{{[}transform Indian agriculture{]}}}}{\small\tt{,he added.}}\end{tabular}                                                                                                                                                       &                                                 \\ \hline
\end{tabular}}
\caption{Examples of tasks mentioned in various datasets. The task mentions are highlighted in bold in square brackets. Each task belongs to an Industry Group.}
\label{tab: exampletask}
\end{table*}

We define the problem as follows. Consider a set of documents $D$ % = \{d_1, d_2, \dots, d_n\}$ %that consists of news or general activities 
and a set of IGs $G=\{g_1, g_2, \dots, g_{24}\}$. %are considered as input. %We define the problem as follows: 
%{\emph{Given the set of IGs, infer common tasks $T = \{t_1, t_2, \dots, t_m\}$ from the documents $D$, such that the tasks $T$ can be directly added to commonsense knowledge bases.}} Making such an end to end framework is not straight forward and requires several level of orchestrations. 
%{\emph{The goal is to discover a set of tasks $T_i = \{t_{i1}, t_{i2}, \dots, t_{im}\}$ for each IG $g_i$, such that the triples of the form $\langle g_i~ company$, \textit{is~capable~of}, task $t_{ij}\rangle$ can be directly added to a commonsense knowledge base. }} %The challenge is to design an end to end framework which is not straightforward and requires several level of orchestrations. 
{\emph{The goal is discover a set of tasks $T_i = \{t_{i1}, t_{i2}, \dots, t_{im}\}$ for each IG $g_i$, such that the triples of the form $\langle g_i~ company$, \textit{is~capable~of}, task $t_{ij}\rangle$ can be directly added to a commonsense knowledge base. }} %The challenge is to design an end to end framework which is not straightforward and requires several level of orchestrations.
In order to achieve this goal, each task extracted from $D$ needs to be {\em matched} with one or more suitable IGs. 
The challenge is to design an end to end weakly supervised framework.

%The first sub-problem is to extract tasks $T = \{t_1, t_2, \dots, t_m\}$ from documents $D$. Next challenge is to assign at most one industry group $g$ to each extracted tasks $T$. Tasks in the documents are not in standard form and requires to be converted in the canonical form. Also, to measure the relevance of a task for a IG $g$, we need to compute the affinity for each task-IG pair. At the end, we choose the top-tasks with highest affinity to be added in the commonsense knowledge base. We show examples of a different set of tasks and the corresponding industry group in Appendix Table~\ref{tab: exampletask}.
%Each of the tasks is enclosed in square brackets and are highlighted in bold within a sentence. For simplicity, we only show a subset of tasks from a document. Note that, each of the tasks corresponds to an industry group, as shown in the table~\ref{tab: exampletask}. The detailed examples of tasks corresponding to the industry group {\small\tt{transportation}} is presented in appendix (ref table~\ref{tab: exampletaskIG})

The specific contributions of this paper are:\\
%\begin{itemize}
%\item 
$\bullet$ A general method to extract domain-specific knowledge from text that can augment knowledge-bases like ConceptNet (Section~\ref{secMethod}) \\
%\item 
$\bullet$ Unsupervised identification of appropriate IGs for tasks extracted from a large corpus (Sections~\ref{secTaskExtraction} and~\ref{secTaskIGClassification}) \\
%\item Self-supervised neural affinity model which learns to predict affinity between a task and an IG taking into consideration corpus-level information (Section~\ref{secAffinityModel})
%\item 
$\bullet$ Self-supervised neural affinity model for matching tasks with IGs, which learns to predict affinity between a task and an IG taking into consideration corpus-level information (Section~\ref{secAffinityModel}) \\
%\item 
$\bullet$ Detailed experimental analysis and evaluation (Section~\ref{secExp})
%\end{itemize}

%\section{Method}
\section{Proposed Method}\label{secMethod}

\begin{figure*}[t]
    \centering
   \includegraphics[width=0.95\linewidth,height=0.15\linewidth]{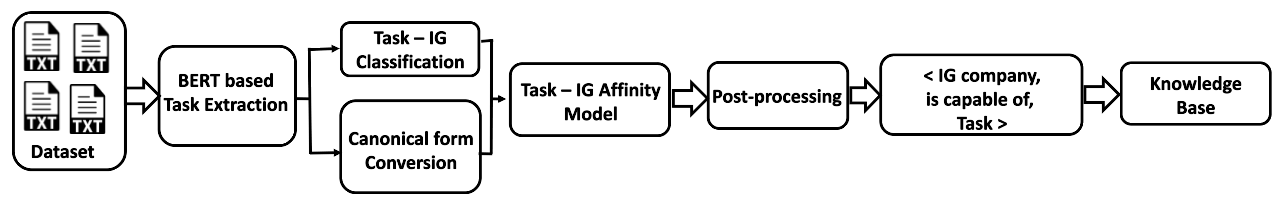}
   %\vspace{-0.1in}
    \caption{End-to-end framework to augment Task-IG information in KB}
    \label{fig:framework}
\end{figure*}

We propose a multi-step technique as shown in Figure~\ref{fig:framework} and the key steps are described below.

%\noindent\textbf{S1: Task Extraction.} 
\subsection{Task Extraction}\label{secTaskExtraction}
In this step, we extract a set of tasks $T$ from the documents $D$ using the task extraction technique proposed by Pawar et al.~\cite{pawar2021weakly}. % with some minor revisions. 
The technique is based on a weakly supervised BERT-based classification model which predicts for each word whether it is the headword of a task phrase or not. %This model is trained using weak supervision provided through Snorkel framework~\cite{ratner2017snorkel}. 
A set of dependency tree based rules are then used to expand a task headword to a complete task phrase. %We enhanced the resources such as list of action nouns and verbs and revised the phrase expansion rules to exclude less informative prepositional phrases. % which only give time or space information about a task.

\begin{algorithm}[h]
\small{
 \KwData{Task ($t$), Set of IGs ($G$), Set of keywords for each IG ($KW=\{KW_g : g\in G\}$), Set of hypotheses ($HP=\{HP_g : g\in G\}$)}
 \KwResult{Task $t$ labelled with $g\in G$ }
 %sentence-bert-model = {\small{\texttt{all-MiniLM-L6-v2}}} \\
 %label\_embedding = embedding($g$, sentence-bert-model)\\
 %\For{$t\in T$}{
  $g_{ks}$ =  keywords\_lookup($t$, $KW$)\\
  %task\_embedding = embedding($t$, sentence-bert-model)\\
  $g_{cs}$, $g'_{cs}$, $g''_{cs}$ =  cosine\_sim($t$, $HP$)  \tcp{\scriptsize returning top 3 most probable labels}
  \lIf{$g_{ks}$ == $g_{cs}$}{
   \KwRet $g_{ks}$
   }
\Else{
   %Select top three labels by $cosine\_similarity$\\
   $g_{zs}$, conf = zero\_shot\_tc($t$, $HP$, $g_{cs}$, $g'_{cs}$, $g''_{cs}$) \tcp{\scriptsize checking only 3 most probable labels for better efficiency}
   \lIf{conf > 0.95 OR $g_{zs}$ == $g_{ks}$}{
   \KwRet $g_{zs}$
   }
   %\Else{
   %  \lIf{$g_{zs}$ == $g_{ks}$}{
   %\KwRet $g_{zs}$\;
   %}
    \lElse{
   \KwRet {\small\sf Others}
   }
   %} 
  }
 %}
}
 \caption{Task-IG classification algorithm}
\label{algTaskIGClassification}
\end{algorithm}
%\noindent\textbf{S2: Task-IG Classification.}
\subsection{Task-IG Classification}\label{secTaskIGClassification}
Next, given the set of extracted tasks $T$, we label each task $t$ to a corresponding IG $g$, i.e., $t \rightarrow g$. Note that some tasks are general and cannot be associated with any IG and are labelled as {\small\sf Others}. Due to unavailability of any labelled training data, we focus on unsupervised methods where the only manual effort is to provide a set of five keywords $kw_i$ associated with each IG $g$ along with a one-sentence description $HP_g$ referred as hypothesis. $HP_g$ is of the form of {\small\tt The previous sentence is about some aspects of $g$ such as $kw_1$ or \dots, $kw_5$} (see Table~\ref{tab: appendix_IG_keywords} for such keywords for each IG). For example, for the IG {\small\sf{Real Estate}}, the following hypothesis is formed using the five keywords -- {\small\texttt{The previous sentence is about some aspects of real estate such as rent or house or residential or apartment or homeowner.}} We use the following three unsupervised methods:\\
%\begin{enumerate}
%\item 
\textbf{1. keywords\_lookup}: %A naive approach to look-up the keywords for each IG to assign a label ($g_{ks}$) to a task $t$.
A naive approach which simply checks for the presence of the keywords for each IG in a task phrase $t$ to assign the corresponding label ($g_{ks}$) to $t$.\\
%\item 
\textbf{2. cosine\_sim}: Cosine similarity scores between embeddings of a task $t$ and the hypothesis $HP_g$ for each IG are computed. The embeddings are obtained using Sentence-BERT~\cite{reimers-2019-sentence-bert} model {\small\tt all-MiniLM-L6-v2}. The resultant IG with maximum cosine similarity is denoted as $g_{cs}$.\\
%\item 
\textbf{3. zero\_shot\_tc}: A zero-shot text classification~\cite{yin2019benchmarking} technique based on Natural Language Inference (NLI) approach. Here, one of the following relations is predicted for a pair of a task $t$ as a {\bf premise} and an $HP_g$ (for each IG) as a {\bf hypothesis} -- {\small\sf Entail} and {\small\sf Contradict}.
%Here, a relation between a task $t$ and an IG $g$ (represented using $HP_g$) is predicted -- {\small\sf Entail} vs {\small\sf Contradict}. 
IG with the highest entailment probability is returned as $g_{zs}$. %with a confidence value. 
We use the pre-trained NLI model {\small\tt distilbart-mnli-12-3}\footnote{\url{https://huggingface.co/valhalla/distilbart-mnli-12-3}} for these predictions.
%\end{enumerate}

We then use a simple ensemble approach to combine predictions of these three methods as described in Algorithm~\ref{algTaskIGClassification} to predict a {\em task-level} IG for a task $t$. We also predict a {\em sentence-level} IG for each task by using the same algorithm to classify the entire sentence containing the task (as against classifying only the task phrase to predict {\em task-level} IG) . The {\em sentence-level} IG for a task gets influenced by other context words in the sentence outside the task phrase itself and hence may be different than the {\em task-level} IG.
%We use the same procedure to also label each {\em sentence} containing at least one task with an appropriate IG.

\begin{table*}[t]\center
\scalebox{0.85}{
\begin{tabular}{|l|l|}
\hline
\multicolumn{1}{|c|}{\textbf{Industry Group}}                                        & \multicolumn{1}{c|}{\textbf{Keywords}}                                                                                                                                                                    \\ \hline
Real Estate                                                                          & rent, house, residential, apartment, homeowner                                                                                 \\ \hline
Utilities                                                                            & plumbing, sewage, bill, housekeeping, laundry                                                                                      \\ \hline
\begin{tabular}[c]{@{}l@{}}Media \& \\ Entertainment\end{tabular}                    & \begin{tabular}[c]{@{}l@{}}news, film, advertising, publishing, broadcasting\end{tabular}                         \\ \hline
Telecommunication                                                                    & \begin{tabular}[c]{@{}l@{}}telephone, mobile, network, internet, wireless communication\end{tabular}           \\ \hline
\begin{tabular}[c]{@{}l@{}}Semiconductors \& \\ Semiconductor equipment\end{tabular} & \begin{tabular}[c]{@{}l@{}}chip, ram, processor, motherboard, cpu\end{tabular}                 \\ \hline
Software \& Services                                                                 & data, app, outsourcing, programming, server                                                                            \\ \hline
\begin{tabular}[c]{@{}l@{}}Diversified \\ Financials\end{tabular}                    & \begin{tabular}[c]{@{}l@{}}investment, stock, portfolio, capital, asset management\end{tabular}                   \\ \hline
\begin{tabular}[c]{@{}l@{}}Health care equipment \\ \& services\end{tabular}         & \begin{tabular}[c]{@{}l@{}}hospital, medical, doctor, nurse, diagnostics\end{tabular}                  \\ \hline
Food \& Staples Retailing                                                            & agriculture, farm, crop, vegetable, fruit                                                                         \\ \hline
\begin{tabular}[c]{@{}l@{}}Technology hardware\\  \& equipment\end{tabular}          & \begin{tabular}[c]{@{}l@{}}gadget, smart phone, tablet, graphic card, storage\end{tabular}              \\ \hline
Insurance                                                                            & \begin{tabular}[c]{@{}l@{}}health insurance, life insurance, medical insurance, risk insurance, insurance brokers\end{tabular} \\ \hline
Banks                                                                                & loan, mortgage, accounts, payment, money                                                                                              \\ \hline
Pharmaceuticals                                                                      & medicine, drug, vaccine, syrup, biotechnology                                                                               \\ \hline
\begin{tabular}[c]{@{}l@{}}Household \& personal\\  products\end{tabular}            & \begin{tabular}[c]{@{}l@{}}toiletry, eyewear, cleansing, cosmetic, beauty product\end{tabular}            \\ \hline
\begin{tabular}[c]{@{}l@{}}Food beverage \&\\  tobacco\end{tabular}                  & \begin{tabular}[c]{@{}l@{}}alcohol, meat, brewer, distillery, cigarette\end{tabular}                            \\ \hline
Retailing                                                                            & \begin{tabular}[c]{@{}l@{}}ecommerce, merchandise, distributor, shop, supermarket\end{tabular}                                 \\ \hline
Consumer Services                                                                    & \begin{tabular}[c]{@{}l@{}}hotel, restaurant, education, resort, casino.\end{tabular}                                   \\ \hline
\begin{tabular}[c]{@{}l@{}}Consumer durables \& \\ apparel\end{tabular}              & \begin{tabular}[c]{@{}l@{}}textile, footwear, electronic appliance, clothing, houseware\end{tabular}        \\ \hline
Automobiles \& components                                                            & \begin{tabular}[c]{@{}l@{}}car, truck, vehicle, motorcycle, tire\end{tabular}                                   \\ \hline
Transportation                                                                       & railway, highway, airline, shipping, logistics                                                                               \\ \hline
\begin{tabular}[c]{@{}l@{}}Commercial \& professional \\ services\end{tabular}       & \begin{tabular}[c]{@{}l@{}}consulting, hiring, human resource, recruitment, printing\end{tabular}    \\ \hline
Capital goods                                                                        & \begin{tabular}[c]{@{}l@{}}machinery, equipment, aerospace, defense, satellites.\end{tabular}                               \\ \hline
Materials                                                                            & metal, mining, fertilizer, chemical, cement                                                                                       \\ \hline
Energy                                                                               & oil, electricity, coal, renewable, solar                                                                                             \\ \hline
\end{tabular}}
\caption{Keywords for 24 Industry Groups (IGs)}
\label{tab: appendix_IG_keywords}
\end{table*}

%\noindent\textbf{S3: Canonical form Conversion.}
\subsection{Canonical form Conversion}
Any task $t$ extracted directly from documents is very specific and needs to be generalized to a {\em canonical form} $\hat{t}$ before adding to a common sense KB. %The canonical form $\hat{t}$ of a task  $t$ represents the standard form of the task. 
For example, consider the task $t$: {\small\tt created virtual creatures and characters in movies like ``The Lord of the Rings''} belonging to the IG {\small\sf Media~\&~Entertainment}. It is very specific to a particular movie and needs to be generalized to say {\small\tt create virtual creatures and characters}.
%We apply each of the above-mentioned rules to convert task $t$ to its canonical form $\hat{t}$. For instance, the above task $t$ transformed to its canonical form $\hat{t}$ is represented as \texttt{create virtual creatures and characters}.\\ 

We use a set of linguistic rules for the conversion which we now describe in brief. % detailed in Appendix~\ref{secCanonical}.
%We devise the following rules to convert a task to its canonical form, i.e., $t \rightarrow \hat{t}$. 
We first remove any noise such as hyphen, quotes, text in brackets, from the task phrase. Any task can either be a verb phrase or a noun phrase. If a task is a verb phrase, we first convert any passive voice phrases to active voice and then identify a syntactic pattern of a task as V-NP (verb followed by a noun phrase), V-NP-P-NP (verb followed by a noun phrase followed by a prepositional phrase), etc. The head verbs are converted to base forms (e.g., {\small\tt analyzed} $\Rightarrow$ {\small\tt analyze}) and the constituent noun phrases (NPs) are processed to remove any determiners (e.g., {\small\tt the}, {\small\tt a}), possessive pronouns (e.g., {\small\tt his}, {\small\tt its}) and common uninformative adjectives (determined using corpus statistics, e.g., {\small\tt some}, {\small\tt only}). %Using a dependency tree, 
We also process presence of any named entities in these constituent NPs, e.g., GPE (geo-political entity) mentions are replaced with {\small\tt state}, {\small\tt country}, or {\small\tt continent} using the pre-defined list and rest as ``places''. Entity mentions of type MONEY, DATE or TIME are removed from the task. The tasks which are noun phrases are similarly processed by first identifying a suitable pattern (such as only NP, NP-P-NP) and then applying the similar rules as above.

%\noindent\textbf{S4: Task-IG Affinity model.}
\subsection{Task-IG Affinity model}\label{secAffinityModel}
%To augment task-IG pair to common sense knowledge base
For addition in a commonsense KB, it is desirable to add most representative tasks of each IG. There may be some tasks that are too specific to an IG while other tasks may be too general. Hence, given a task in its canonical form $\hat{t}$ labelled with IG $g$, we devise a function $f(\hat{t},g)$ which predicts a affinity score $\in [-1, 1]$. The function is expected to return a high affinity score if the following conditions hold: \\
\noindent\textbf{High support:} The task $\hat{t}$ (or other tasks with {\em similar} meaning) is observed with IG $g$ in multiple sentences in the corpus.\\
\noindent\textbf{High specificity:} The task $\hat{t}$ (or other tasks with {\em similar} meaning) is specific to $g$, i.e., it is rarely observed for any IGs other than $g$ in the corpus.\\
%The Task-IG Affinity model is motivated from the word2vec skipgram~\cite{mikolov2013efficient} model where the goal is to learn word embeddings for words such that the words having similar meaning (i.e., occurring in similar context) would be assigned word embeddings which are closer to each other. %The idea is to learn a binary classifier in a self-supervised way. 
%In our context, the tasks having similar meaning, observed with an IG needs to be grouped together. We create the training data automatically by sampling task-IG pair. The positive and negative instances are as follows: 
To learn such an affinity function $f(\hat{t},g)$, we use self-supervision where the training instances are created as follows for each task $\hat{t}$ labelled with an IG $g$: $\langle \hat{t}, g, g_{n_1}, g_{n_2}, w_i\rangle$. 
%\begin{equation*}\scriptsize
%\begin{aligned}
%\langle t, g, g_{n_1}, g_{n_2}\rangle
%\end{aligned}
%\end{equation*}
%\noindent\textbf{Positive instances:} For $k$ tasks labelled with IG $g$, we create one positive instances per task-IG pair $ \langle t_1,g \rangle,\langle t_2,g \rangle,…, \langle t_k,g\rangle$
%\noindent\textbf{Negative instances:} For each positive instance, we randomly select any two IGs ($g'$ and $g''$) other than $(g, g_{cs_{1}}, g_{cs_{1}}, g_{cs_{2}})$ to create $2-k$ negative instances - $\langle t_1,g'\rangle,\langle t_2,g'\rangle, \dots \langle t_k,g'\rangle,\linebreak\langle t_1,g''\rangle, \langle t_2,g'' \rangle, \dots, \langle t_k,g'' \rangle$.
Here, $g$ is either {\em task-level} or {\em sentence-level} IG predicted for task $t$ and $w_i$ is the instance weight that is described later. $g_{n_1}$ and $g_{n_2}$ are negative IGs for $\hat{t}$ which are randomly selected from IGs other than $g$ and $G_{EX}$, where $G_{EX} = \{g_{cs}, g'_{cs}, g''_{cs}\}$ represents the three most probable IGs returned by cosine similarity based technique. Note that excluding the IGs from $G_{EX}$ gives more robust negative instances.  %We describe the task-IG affinity model below.

\noindent\textbf{Model Description.} %We encode the positive and negative instances along with the label using a pre-trained sentence-transformers (SM) - \texttt{all-MiniLM-L6-v2}\citep{reimers-2019-sentence-bert}.
We encode the tasks as well as IGs (each IG $g$ is represented using the corresponding hypothesis $HP_g$) using a pre-trained sentence-transformer (SM) -- {\small\tt all-MiniLM-L6-v2}%~\cite{reimers-2019-sentence-bert}
, to obtain task and IG embeddings in $\mathbb{R}^{384}$.
\begin{eqnarray*}\small
x_t = SM(\hat{t}); 
x_g = SM(HP_g)\\
x_{g_{n_1}} = SM(HP_{g_{n_1}}); 
x_{g_{n_2}} = SM(HP_{g_{n_2}})
\end{eqnarray*} 
Then we pass it through two linear feed forward layers to get more compressed representations ($\in \mathbb{R}^{100}$) of the task and IG embeddings. We use $tanh$ activation and a dropout with 0.25 probability. These new representations are also expected to be more informed about the corpus-level information as compared to the original sentence-transformer embeddings, because of the self-supervised training process described further. 
\begin{eqnarray*}
x'_t = tanh(W_t\cdot x_t+ b_t)\\
x'_g = tanh(W_g\cdot x_g+ b_g)\\
x'_{g_{n_1}} = tanh(W_g\cdot x_{g_{n_1}}+ b_g) \\
x'_{g_{n_2}} = tanh(W_g\cdot x_{g_{n_2}}+ b_g)
\end{eqnarray*} 
Here, $W_t, W_g\in \mathbb{R}^{100\times 384}$ and $b_t, b_g\in\mathbb{R}^{100}$ are learnable weights of the linear transformation layers. Using these representations, we compute the cosine similarity between the task and each of the IGs in the instance. %(one positive and two negative IGs). %embeddings for labels and get a compressed representation as $\{g_{pos}, g_{neg_1}, g_{neg_2}\} \in \mathbb{R}^{100}$. Next, we compute the cosine similarity.
\begin{eqnarray}\label{eqSim}
f(\hat{t},g)= sim_{pos} = CosineSim(x'_t, x'_g)\\
sim_{neg_1} = CosineSim(x'_t, x'_{g_{n_1}})\\
sim_{neg_2} = CosineSim(x'_t, x'_{g_{n_2}})
\end{eqnarray} 
%We then compute the margin ranking loss to update weights such that the semantically closer representations are pushed closer to each as compared to the semantically different representation. Here, we use the margin value $\epsilon$ as 0.5 and the loss function is defined as follows.
We then compute the margin ranking loss to update weights such that the representations of semantically similar $\hat{t}$ and $g$ are pushed closer to each other and vice versa. Here, we use the margin value $\epsilon$ as 0.5 and the loss function is defined as follows.
\begin{eqnarray*}
loss_1 = max(0, -(sim_{pos} - sim_{neg_1}) + \epsilon)\\
loss_2 = max(0, -(sim_{pos} - sim_{neg_2}) + \epsilon)\\
%ML(x_1, x_2, \epsilon) = max(0, -(x_1 - x_2) + \epsilon)\\
%loss_1 = ML(similarity_{pos}, similarity_{neg_{1}}, \epsilon)\\
%loss_2 = ML(similarity_{pos}, similarity_{neg_{2}}, \epsilon)\\
loss = w_i\cdot(loss_1 + loss_2)
\end{eqnarray*}

\begin{table*}[t]\center
\scalebox{0.8}{
\begin{tabular}{|cl|cl|}
\hline
%\multicolumn{2}{|l|}{\textbf{\hspace{1.1in} < IG company,~~~~is~capable~of,~~~~    Task >}}                                                                                                                                              & \multicolumn{2}{l|}{\textbf{\hspace{1.1in} < IG company,~~~~is~capable~of,~~~~    Task >}}                                                                                           \\ \hline
%\multicolumn{1}{|c|}{\sf Real Estate}                                                                         & \begin{tabular}[c]{@{}l@{}}-{\small\tt save landlords time and money} \\ -{\small\tt represent rental units}\end{tabular}  & \multicolumn{1}{c|}{\sf Media \& Entertainment} & \begin{tabular}[c]{@{}l@{}}-{\small\tt make money from advertising}\\ -{\small\tt understand news content}\end{tabular}                          \\ \hline
\multicolumn{1}{|c|}{\small\sf Media \& Entertainment} & \begin{tabular}[c]{@{}l@{}}-{\small\tt make money from advertising}\\ -{\small\tt understand news content}\end{tabular} & \multicolumn{1}{|c|}{\small\sf Real Estate}                                                                         & \begin{tabular}[c]{@{}l@{}}-{\small\tt save landlords time and money} \\ -{\small\tt represent rental units}\end{tabular}\\ \hline
\multicolumn{1}{|c|}{\small\sf Telecommunication Services}                                                          & \begin{tabular}[c]{@{}l@{}}-{\small\tt provide wi-fi access for devices}\\ -{\small\tt cellular telephone operations}\end{tabular}           & \multicolumn{1}{c|}{\small\sf Banks}                  & \begin{tabular}[c]{@{}l@{}}-{\small\tt manage payment services}\\ -{\small\tt lend overnight money}\end{tabular}                                        \\ \hline
\multicolumn{1}{|c|}{\begin{tabular}[c]{@{}c@{}}{\small\sf Semiconductors} \&\\ {\small\sf Semiconductor Equipment}\end{tabular}} & \begin{tabular}[c]{@{}l@{}}-{\small\tt chip designs for autonomous systems} \\ -{\small\tt build semiconductor foundries}\end{tabular} & \multicolumn{1}{c|}{\small\sf Insurance}              & \begin{tabular}[c]{@{}l@{}}-{\small\tt sell insurance products} \\ -{\small\tt get insurance license}\end{tabular}                           \\ \hline
%\multicolumn{1}{|c|}{\sf Diversified Financials}                                                              & \begin{tabular}[c]{@{}l@{}}{\small\tt make targeted investments,} \\ {\small\tt lead investments in companies}\end{tabular}           & \multicolumn{1}{c|}{\sf Software \& Services}   & \begin{tabular}[c]{@{}l@{}}{\small\tt build smartphone apps,} \\ {\small\tt analysis of apps and data transmissions}\end{tabular}               \\ \hline
\multicolumn{1}{|c|}{\small\sf Pharmaceuticals}                                                                     & \begin{tabular}[c]{@{}l@{}}-{\small\tt fund vaccine trials}\\ -{\small\tt develop therapeutic medicines}\end{tabular}                & \multicolumn{1}{c|}{\small\sf Retailing}              & \begin{tabular}[c]{@{}l@{}}-{\small\tt launch e-commerce stores}\\-{\small\tt deliver grocery orders to customers}\end{tabular} \\ \hline
\end{tabular}}
%\vspace{-0.1in}
\caption{Task-IG pairs to be added to a commonsense KB in the form of $\langle${\sf IG company}, is~capable~of, {\tt Task}$\rangle$}
\label{tabTaskIGEx}
\end{table*}

%\begin{table}\scriptsize
%\begin{tabular}{p{0.38\columnwidth}p{0.57\columnwidth}}
%\hline
%{\sf Media \&} & -{\tt make money from advertising} \\
%{\sf Entertainment} & -{\tt understand news content} \\
%{\sf Real Estate} & -{\tt save landlords time and money} \\
% & -{\tt represent rental units}\\
%\hline
%{\sf Telecommunication} & -{\tt provide wi-fi access for devices} \\
%{\sf Services} & -{\tt cellular telephone operations} \\
%\hline
%{\sf Semiconductors \&} & -{\tt chip designs for autonomous systems} \\
%{\sf Semiconductor Equipment} & -{\tt build semiconductor foundries} \\
%\hline
%{\sf Pharmaceuticals} & -{\tt fund vaccine trials} \\
% & -{\tt develop therapeutic medicines} \\
%\hline
%\end{tabular}
%\caption{Task-IG pairs to be added to a commonsense KB in the form of $\langle${\sf IG company}, is~capable~of, {\tt Task}$\rangle$}
%\label{tabTaskIGEx}
%\end{table}
\begin{figure*}\center
\includegraphics[width=0.95\linewidth,height=0.4\linewidth]{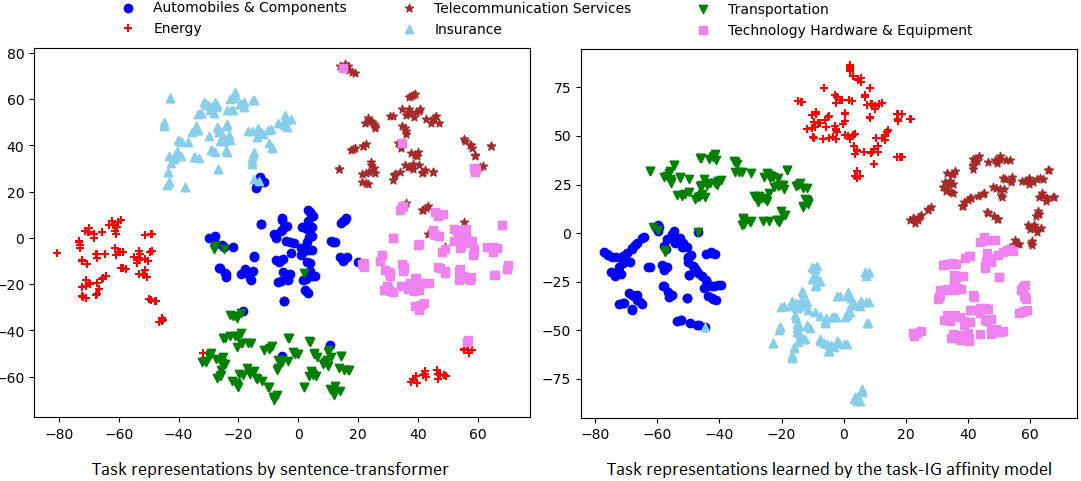}
\caption{Comparative representations of tasks}
\label{figTaskRepr}
\end{figure*}
        
\noindent\textbf{Instance weighting.} Three different types of instance weights are considered. \\
%The training set contains significantly imbalanced distribution of examples among the various IGs, e.g., 22958 tasks labelled with {\small\sf Software \& Services} compared to only 8994 of {\small\sf Banks}. %Also, as the labels are obtained based on some sequence of sentence (task), the context based on sentence is lost. It may add noise to the labelled task-IG pair. 
%To alleviate this problem, we assign a weight ($w_{i}$) to each instance such that total weight for all instances in each IG $g$ is equal across all IGs. 
(i) \textbf{Balancing weight} ($w_b$) is assigned to each instance to handle imbalanced distribution of instances across IGs (e.g., 22958 tasks labelled with {\small\sf Software \& Services} compared to only 8994 of {\small\sf Banks}). $w_b$ is calculated as $\frac{N}{24\cdot N_g}$ where $N$ is the total number of training instances and $N_g$ is the number of instances with IG $g$. \\
%such that total weight for all instances in each IG $g$ is equal across all IGs. 
%We use sample weighting scheme to alleviate the problem. We weight each sample $t$ tagged with IG $g$ as follows:
%\begin{equation*}\small
%w_{iw} = \frac{Total~training~samples * || IG||}{Total~training~samples~for~class~g}
%\end{equation*}
%The weighting scheme gives higher weights to the instances containing $g$ which is present in less number of instances and vice versa. % samples with less no of IG $g$ and gives lower weight for the classes with more examples. %It helps the model to learn robustly across all classes and solves the problem of skewed learning. 
(ii) \textbf{Confidence weight} ($w_{c}$) captures the confidence of the predicted IG $g$ as follows -- (a) $w_{c}=1$, if both {\em task-level} and {\em sentence-level} IGs are same as $g$, (b) $w_{c}=0.75$, if {\em task-level} IG is $g$ but {\em sentence-level} IG is different, (c) $w_{c}=0.25$, if {\em sentence-level} IG is $g$ and {\em task-level} IG is {\small\sf Others}. \\
%Another instance weight ($w_{c}$) is considered to capture the confidence of predicted IG as follows -- (i) 1.0 if both {\em task-level} and {\em sentence-level} IGs are same as $g$, (ii) 0.75 if {\em task-level} IG is $g$ but {\em senence-level} IG is different, (iii) 0.25 if {\em sentence-level} IG is $g$ and {\em task-level} is {\small\sf Others}.  
%In addition, we address the noisy label prediction problem as follows. If the task level prediction and sentence level prediction are same, we give a higher weight ($w_{pt}$) of $1$. If the task level prediction is $g$ (except {\small\sf Others}) and sentence level prediction is different than $g$ then we assign a weight ($w_{pt}$) of $0.75$. If the task level prediction is {\small\sf Others} and sentence level prediction is different IG $g$ then we change the original task prediction to $g$ with the weight $w_{pt}$ as $0.25$. %The weighing scheme $w_{pt}$ helps the model to overcome the problem of noise labels. 
%The weighing scheme $w_{pt}$ gives more weight to the tasks with higher confidence in the labels and vice versa. 
%To make the sample weighing more robust and comprehensive, we also extract {\em agent} for each task using dependency relations $nsubj$ and $agent$. Note that the tasks with an agent as some organization are more likely to have correct IG. We assign the weight $w_{at}$ as $1.0$ for such tasks and set $w_{at}$ as $0.5$ for other tasks. %if no agent is observed.
(iii) \textbf{Agent-based weight} ($w_{a}$) assigns a higher weight of 1 to tasks which have some organization as their {\em agent} in the sentence and lower weight of $0.5$ to other tasks. \\
%is considered to assign higher weight of 1 to tasks which have some organization as their {\em agent} in the sentence and lower weight of $0.5$ to other tasks. 
%Additionally, another instance weight ($w_{a}$) is considered to assign higher weight of 1 to tasks which have some organization as their {\em agent} and lower weight of $0.5$ to other tasks. 
%\vspace{-0.1in}
%\begin{equation*}\small
%\begin{aligned}
%weight = w_{iw} \times w_{pt} \times w_{at}; 
%loss = loss \times weight
%\end{aligned}
%\end{equation*}
Finally, the overall weight of an instance is $w_i=w_b\cdot w_{c}\cdot w_{a}$ which is multiplied with the loss.

\noindent\textbf{Training.} %We train the model with batch size of $64$ samples using Adam optimizer with a learning rate of $0.0001$. With the pre-trained embeddings freezed, we train the model for $5$ epochs. The model returns the cosine similarity between any task-IG pair as the affinity score. 
We train the model with the following settings on a {\small\tt Intel(R) Xeon(R) Gold 6148@2.40GHZ} machine with 48 {\small GB RAM}: batch size of $64$, Adam optimizer with learning rate of $0.0001$, and $5$ epochs. 
During {\em inference}, for a task-IG pair $\langle \hat{t}, g\rangle$, the affinity score $f(\hat{t},g)$ is computed using the learned model as $sim_{pos}$ (Eq.~\ref{eqSim}).
%The model returns the cosine similarity between any task-IG pair as the affinity score. 

%Here, each word is represented by two vectors – target word vector () and context word vector (). The vectors for all words in the vocabulary are initialized randomly. During the model training, for a positive pair , the word vectors are updated such that the dot product  is increased and for a negative pair , the word vectors are updated such that the dot product  is decreased. After the model the trained using the positive and negative instances created from a large corpus, the learned word vectors (either  or ) are used as word2vec word embeddings for the word .

%\noindent\textbf{S5: Post-processing}. %At the end, we cluster the tasks and return the top $k$ tasks per IG to be added to a knowledge base.
%We cluster the tasks within each IG and retain only one task per cluster. Table~\ref{tab: task_ig_affinity_example} illustrates examples of tasks added for some IGs.
\subsection{Post-processing}
We cluster the tasks within each IG using community detection algorithm provided by the sentence-transformers package\footnote{https://www.sbert.net/}. A task with the highest affinity score is selected from each cluster as its representative. Table~\ref{tabTaskIGEx} illustrates examples of tasks added for some IGs. Figure~\ref{figTaskRepr} shows visualization of task representations learned by the affinity model for some IGs, as compared with the original sentence-transformer representations. It can be observed that the task representations learned by the affinity model provide relatively better separation.
%We cluster the tasks within each IG using task embeddings obtained from the sentence Transformer and retain only one task per cluster. Table~\ref{tab: task_ig_affinity_example} illustrates examples of tasks added for some IGs.

%\noindent\textbf{Summarise End to End} //TODO

%\section{Results}
\section{Experiments}\label{secExp}

\subsection{Dataset}
%\textbf{Dataset.} 
We use two publicly available news datasets: TechCrunch\footnote{https://www.kaggle.com/sumantindurkhya/techarticles2020} (consists of technical news published on TechCrunch in 2020) and Reuters~\cite{lewis1997reuters} which is a collection of documents from the well known dataset from financial newswire service. %We briefly provide the details of the dataset in Table~\ref{tab: dataset description}. 
See Table~\ref{tab: dataset description} for details. 
%The experiments were carried out on a {\small\tt Intel(R) Xeon(R) Gold 6148@2.40GHZ} machine with 48 GB RAM.
%\vspace{\medskipamount}
\begin{table}[]\center

\scalebox{0.75}{
\begin{tabular}{cccc}
\hline
%\textbf{Dataset}    & \textbf{\begin{tabular}[c]{@{}c@{}}Number of \\ News Item\end{tabular}} & \textbf{\begin{tabular}[c]{@{}c@{}}Number of \\ Sentences\end{tabular}} & \textbf{\begin{tabular}[c]{@{}c@{}}Number of \\ Task Extracted\end{tabular}} \\ \hline
\textbf{Dataset}    & \textbf{\#News items} & \textbf{\#Sentences} & \textbf{\#Tasks Extracted} \\ \hline
\textbf{TechCrunch} & 19081                                                                   & 458268                                                                  & 450071                                                                       \\ %\hline
\textbf{Reuters}    & 10518                                                                   & 60034                                                                   & 43466                                                                        \\ \hline
\end{tabular}}
%\vspace{-0.1in}
\caption{Dataset Description}
\label{tab: dataset description}
\end{table}
%\noindent{\textbf{Evaluation of Tasks Extraction.}} We do an extensive evaluation of each phase of the framework. As no public gold dataset is available, we manually annotate the dataset. In order to create the ground truth for evaluating task extraction technique, we manually annotated 30 files from TechCrunch and Reuters with gold standard task phrases\footnote{Evaluation datasets will be made available upon request.}. The evaluation dataset consists of a total of 890 tasks. On an average, we get the precision of 0.82 and recall of 0.72 on this dataset.

\subsection{Evaluation}
Each phase of the framework is evaluated extensively\footnote{Evaluation datasets will be made available upon request.}.

\noindent{\textbf{Evaluation of Tasks Extraction.}} We manually annotated 30 files from TechCrunch and Reuters with 890 gold standard task phrases. The precision of 0.82 and recall of 0.72 is observed on this dataset.

%\noindent{\textbf{Evaluation of Task-IG Classification.}} We randomly selected 50 tasks labelled with each IG and manually verified the precision. On an average, we observed the precision of 0.76 across all IGs and for 14 IGs, the precision is more than 0.8. Refer Appendix Table~\ref{tab: appendix_result_IG_classification} for the detailed results.
\noindent{\textbf{Evaluation of Task-IG Classification.}} We randomly selected 50 tasks labelled with each IG and %manually verified the precision. 
verified manually. We observed the precision of 0.76 across all IGs and for 14 IGs, the precision is more than 0.8 %Refer Appendix Table~\ref{tab: appendix_result_IG_classification} for the detailed results.
(details in Table~\ref{tab: appendix_result_IG_classification}).

\begin{table*}[t]\center
\scalebox{0.71}{
\begin{tabular}{|l|c|l|c|l|c|}
\hline
\multicolumn{1}{|c|}{\textbf{Industry Group}}                                & \textbf{Precision} & \multicolumn{1}{c|}{\textbf{Industry Group}}                                         & \textbf{Precision} & \multicolumn{1}{c|}{\textbf{Industry Group}}                                   & \textbf{Precision} \\ \hline
Real Estate                                                                  & 0.74               & \begin{tabular}[c]{@{}l@{}}Semiconductors \& \\ Semiconductor Equipment\end{tabular} & 0.98              & Insurance                                                                      & 0.92               \\ \hline
Utilities                                                                    & 0.48               & Software \& Services                                                                 & 0.86              & Pharmaceuticals                                                                & 0.76               \\ \hline
Media \& Entertainment                                                       & 0.86               & Diversified Financials                                                               & 0.96              & \begin{tabular}[c]{@{}l@{}}Food \& Staples \\ Retailing\end{tabular}           & 0.92               \\ \hline
Telecommunication Services                                                   & 0.60               & Banks                                                                                & 0.84              & Retailing                                                                      & 0.84               \\ \hline
\begin{tabular}[c]{@{}l@{}}Technology Hardware \\ \& Equipment\end{tabular}  & 0.82               & \begin{tabular}[c]{@{}l@{}}Household \&\\ Personal Products\end{tabular}             & 0.84              & \begin{tabular}[c]{@{}l@{}}Food Beverage \&\\ Tobacco\end{tabular}             & 0.78               \\ \hline
\begin{tabular}[c]{@{}l@{}}Health Care Equipment \\ \& Services\end{tabular} & 0.92               & Consumer Services                                                                    & 0.70              & \begin{tabular}[c]{@{}l@{}}Consumer Durables\\  \& Apparel\end{tabular}        & 0.92               \\ \hline
Automobiles \& Components                                                    & 0.86               & Transportation                                                                       & 0.48              & \begin{tabular}[c]{@{}l@{}}Commercial \& Professional \\ Services\end{tabular} & 0.84               \\ \hline
Capital Goods                                                                & 0.36               & Materials                                                                            & 0.62              & Energy                                                                         & 0.38               \\ \hline
Others                                                                       & 0.80               & Average &    0.76               &                                                                                &                    \\ \hline
\end{tabular}}
\caption{Precision score for task- G classification for 24 IGs.}
\label{tab: appendix_result_IG_classification}
\end{table*}

\noindent\textbf{Evaluation of Task-IG Affinity.} We compare the task-IG affinity model with the following two baselines. The input to the baselines are the extracted canonical tasks and the corresponding IG labels.\\
%\begin{itemize}
%\item 
$\bullet$ \textit{Association Rule Mining (ARM)~\cite{agrawal1993mining} method.} After lemmatization and removal of stopwords and duplicates, for each task-IG pair we form an itemset with tokens as the words in the task and its IG. %If a token appears multiple times within a single task mention then only one occurrence is retained. %in the corresponding itemset. 
%Now, we get a set of Itemsets $IS = \{I_1, I_2, \dots, I_m\}$ where each itemset is a set of tokens and must contain a IG $g$.
Next, we apply apriori rule mining technique\footnote{https://pypi.org/project/apyori/} to derive the association rules. We retain only those rules having one or more IGs on the right side. Each task-IG pair is assigned a score proportional to the number of matching association rules and their confidence values, and the top 100 tasks for each IG are selected.\\
%To compare with the task-affinity model, we select the top 100 tasks for each IG based on the number of rules and the confidence values. \\
%\textit{TF-IDF method.} For each IG $g$, we create a pseudo document by concatenating all the tasks labeled by that IG to get the term frequency $TF$ of each word. Document frequency $DF$ of a word is computed as the number of distinct IGs in which the word appears. Here, for a word, the method returns one TF-IDF score for each IG. Next, for each task, we compute the affinity score with an IG as the mean value of TF-IDF weights of the words. We sort the affinity scores to get the top 100 tasks for each IG. Note that, the method tries to satisfy the high support and high specificity conditions. %and is a strong baseline.
%\item 
$\bullet$ \textit{TF-IDF method.} For each IG $g$, we create a pseudo document $D_g$ by concatenating all the tasks labeled with $g$ in the corpus. For each word $w$, we compute its term frequency $TF_g(w)$ as the number of times $w$ appears in $D_g$. Its document frequency $DF(w)$ is computed as the $|\{D_g| w\in D_g, \forall_g\}|$. Here, the affinity between a word $w$ and IG $g$ is computed as TF-IDF score $= TF_g(w)\cdot log(\frac{24}{DF(w)})$. Next, for each task, we compute the affinity score with an IG as the mean value of TF-IDF scores of its constituent words. We sort the affinity scores to get the top 100 tasks for each IG. This method tries to satisfy both the desired conditions -- high support (through TF) and high specificity (through IDF). %and is a strong baseline.
%\end{itemize}

\subsection{Results}
%No separate evaluation dataset is created for evaluating task-IG affinity scores, because it is a {\em corpus-level} problem. 
No separate test dataset is required for evaluating task-IG affinity scores, because the problem is to discover tasks with high affinity to each IG from the entire corpus. 
To evaluate the task-IG affinity scores by our technique as well as the above baselines, we compute the metric {\em P@100}. It is computed by manually verifying the precision within top 100 tasks per IG with the highest affinity score for that IG. In other words, approx. 2400 tasks are verified manually (100 tasks per IG, 24 IGs)\footnote{All the discovered Task-IG triples will be made available upon request.} for each of the 3 techniques shown in Table~\ref{tab: appendix_result_task_IG_affinity}. 
%To evaluate the task-IG affinity scores by our technique as well as the above baselines, we use the precision within top 100 tasks per IG. %We consider the precision as the evaluation metric. 
%Table~\ref{tab: pr_score_task_IG} shows the comparative performance of all the techniques. Observe that 
The proposed task-IG affinity model outperforms both the baselines and the differences are {\em statistically significant as per two-sample t-test}.  
\begin{table*}[t]\center
\scalebox{0.55}{
\begin{tabular}{|l|c|c|c|l|c|c|c|l|c|c|c|}
\hline
\multicolumn{1}{|c|}{\textbf{Industry Group}}                                & \multicolumn{1}{l|}{\textbf{ARM}} & \multicolumn{1}{l|}{\textbf{TF-IDF}} & \textbf{\begin{tabular}[c]{@{}c@{}}Affinity\\ Model\end{tabular}} & \multicolumn{1}{c|}{\textbf{Industry Group}}                                         & \multicolumn{1}{l|}{\textbf{ARM}} & \multicolumn{1}{l|}{\textbf{TF-IDF}} & \textbf{\begin{tabular}[c]{@{}c@{}}Affinity\\ Model\end{tabular}} & \multicolumn{1}{c|}{\textbf{Industry Group}}                                   & \multicolumn{1}{l|}{\textbf{ARM}} & \multicolumn{1}{l|}{\textbf{TF-IDF}} & \textbf{\begin{tabular}[c]{@{}c@{}}Affinity\\ Model\end{tabular}} \\ \hline
Real Estate                                                                  & 0.07                              & 0.62                                 & 0.81                                                              & \begin{tabular}[c]{@{}l@{}}Semiconductors \& \\ Semiconductor Equipment\end{tabular} & 0.74                              & 0.75                                 & 0.95                                                              & Insurance                                                                      & 0.68                              & 0.84                                 & 0.88                                                              \\ \hline
Utilities                                                                    & 0.33                              & 0.37                                 & 0.30                                                              & Software \& Services                                                                 & 0.74                              & 0.71                                 & 0.93                                                              & Pharmaceuticals                                                                & 0.89                              & 0.84                                 & 0.95                                                              \\ \hline
Media \& Entertainment                                                       & 0.87                              & 0.84                                 & 0.97                                                              & Diversified Financials                                                               & 0.81                              & 0.67                                 & 0.94                                                              & \begin{tabular}[c]{@{}l@{}}Food \& Staples \\ Retailing\end{tabular}           & 0.71                              & 0.83                                 & 0.96                                                              \\ \hline
Telecommunication Services                                                   & 0.36                              & 0.72                                 & 0.89                                                              & Banks                                                                                & 0.69                              & 0.74                                 & 0.86                                                              & Retailing                                                                      & 0.88                              & 0.88                                 & 0.92                                                              \\ \hline
\begin{tabular}[c]{@{}l@{}}Technology Hardware \\ \& Equipment\end{tabular}  & 0.33                              & 0.64                                 & 0.86                                                              & \begin{tabular}[c]{@{}l@{}}Household \&\\ Personal Products\end{tabular}             & NA                                & 0.62                                 & 0.41                                                              & \begin{tabular}[c]{@{}l@{}}Food Beverage \&\\ Tobacco\end{tabular}             & 0.93                              & 0.93                                 & 0.88                                                              \\ \hline
\begin{tabular}[c]{@{}l@{}}Health Care Equipment \\ \& Services\end{tabular} & 0.79                              & 0.8                                  & 0.90                                                              & Consumer Services                                                                    & 0.60                              & 0.73                                 & 0.87                                                              & \begin{tabular}[c]{@{}l@{}}Consumer Durables\\  \& Apparel\end{tabular}        & 0.74                              & 0.68                                 & 0.75                                                              \\ \hline
Automobiles \& Components                                                    & 0.82                              & 0.77                                 & 0.94                                                              & Transportation                                                                       & 0.57                              & 0.8                                  & 0.94                                                              & \begin{tabular}[c]{@{}l@{}}Commercial \& Professional \\ Services\end{tabular} & 0.68                              & 0.90                                 & 0.88                                                              \\ \hline
Capital Goods                                                                & 0.52                              & 0.73                                 & 0.86                                                              & Materials                                                                            & 0.76                              & 0.69                                 & 0.91                                                              & Energy                                                                         & 0.70                              & 0.82                                 & 0.87                                                              \\ \hline
\textbf{Micro average for all 24 IGs}                                                          & \textbf{0.66}                     & \textbf{0.75}                        & \textbf{0.86}                                                     &  \textbf{Macro average for all 24 IGs}    & \textbf{0.63}             &   \textbf{0.75}            &      \textbf{0.85}                                                                                                                  &                                                                                & \multicolumn{1}{l|}{}             & \multicolumn{1}{l|}{}                &                                                                   \\ \hline
\end{tabular}}
\caption{Comparative P@100 scores of our proposed task-IG affinity model and the baselines for 24 IGs. %The detailed output of our task-IG Affinity model is shared as part of the supplementary data. 
The difference between the P@100 values of task-IG affinity model and both the baselines is statistically significant. Two-sample t-test was carried out considering 24 samples (per IG) leading to the p-values less than 0.01 for both baselines. (The small change in micro and macro average is due to less than 100 tasks in case of a few rare IGs.)}
\label{tab: appendix_result_task_IG_affinity}
\end{table*}

\subsection{Comparing with existing Task-IG triples in ConceptNet}\label{secCN}
To check the number of triples of the form $\langle${\small\sf IG company}, {\it is capable of}, {\small\tt task}$\rangle$ already present in ConceptNet 5.7, we first identified all the entities which represent some type of organization by using {\it IsA} relation triples in ConceptNet. We expand the set of organization entities by using {\it Synonym} relation triples. Finally, we identify the subset of {\it is capable of} relation triples from ConceptNet where the left side entity (head) is one of the organization entities. We found 1372 such triples which were manually verified to get the final list of 138 triples. Each of these 138 triples were also classified manually to assign an appropriate IG as shown in Table~\ref{tab: appendix_addition_ConceptNet}. This shows that our proposed technique has extracted significantly more knowledge compared to what is already present in the existing KB ConceptNet (our 2339 triples vs. 138 triples in ConceptNet). %All these 138 triples are also shared as a part of supplementary data.
%This is a section in the appendix.
\begin{table*}[!htbp]\center
\scalebox{0.7}{
\begin{tabular}{|l|c|l|c|l|c|}
\hline
\multicolumn{1}{|c|}{\textbf{Industry Group}}                                & \textbf{\# Triples} & \multicolumn{1}{c|}{\textbf{Industry Group}}                                         & \textbf{\# Triples} & \multicolumn{1}{c|}{\textbf{Industry Group}}                                   & \textbf{\# Triples} \\ \hline
Real Estate                                                                  & 0               & \begin{tabular}[c]{@{}l@{}}Semiconductors \& \\ Semiconductor Equipment\end{tabular} & 0              & Insurance                                                                      & 1              \\ \hline
Utilities                                                                    & 0            & Software \& Services                                                                 & 2              & Pharmaceuticals                                                                & 1              \\ \hline
Media \& Entertainment                                                       & 10               & Diversified Financials                                                               & 2              & \begin{tabular}[c]{@{}l@{}}Food \& Staples \\ Retailing\end{tabular}           & 2               \\ \hline
Telecommunication Services                                                   & 11               & Banks                                                                                & 16              & Retailing                                                                      & 29               \\ \hline
\begin{tabular}[c]{@{}l@{}}Technology Hardware \\ \& Equipment\end{tabular}  & 0               & \begin{tabular}[c]{@{}l@{}}Household \&\\ Personal Products\end{tabular}             & 0              & \begin{tabular}[c]{@{}l@{}}Food Beverage \&\\ Tobacco\end{tabular}             & 8               \\ \hline
\begin{tabular}[c]{@{}l@{}}Health Care Equipment \\ \& Services\end{tabular} & 1               & Consumer Services                                                                    & 45              & \begin{tabular}[c]{@{}l@{}}Consumer Durables\\  \& Apparel\end{tabular}        & 0               \\ \hline
Automobiles \& Components                                                    & 4               & Transportation                                                                       & 0             & \begin{tabular}[c]{@{}l@{}}Commercial \& Professional \\ Services\end{tabular} & 0               \\ \hline
Capital Goods                                                                & 5               & Materials                                                                            & 0              & Energy                                                                         & 0               \\ \hline
\textbf{Total}                                                                       & \textbf{138}                  & & & & \\ \hline
\end{tabular}}
\caption{Number of triples in the form of $\langle${\small\sf IG company}, is~capable~of, {\small\tt Task}$\rangle$ in ConceptNet KB.}
\label{tab: appendix_addition_ConceptNet}
\end{table*}

\subsection{Limitations}
Some key limitations of our proposed method are as follows which we plan to address as future work.\\
%\begin{enumerate}
%\item 
$\bullet$ \textit{Error propagation}: As ours is a multi-step method, the errors in earlier steps propagate to later steps. For example, if an incorrect task is extracted in the task extraction step, it may lead to an incorrect task to be added in ConceptNet for an IG, unless it is labelled as {\small\sf Others} in the task-IG classification step.\\
%\item 
$\bullet$ \textit{Focussed on a single relation type}: There are multiple relation types in ConceptNet (or in any commonsense KB) but the current work focusses on only one of the relation types, i.e., {\it is capable of}.\\
%\item 
$\bullet$ \textit{No explicit evaluation for recall}: It is not possible to estimate true recall of our method as there is no fixed gold-standard set of tasks that may be carried out by each IG (and also present within the corpus under consideration). Hence, we only estimate precision within top-k tasks identified for each IG. \\
%\item 
$\bullet$ \textit{No extensive hyperparameter search}: Due to unavailability of gold-standard labelled data, it is not possible to carry out extensive hyperparameter search. Because our only evaluation metric (P@100) is the precision in top-k predictions (top-k tasks per IG as per affinity score) which needs significant human annotation (verification) efforts. For each hyperparameter setting, a different set of tasks would be part of the top-k and needs to be validated by human annotators separately. %Due to these restrictions
Hence, we could not carry out extensive hyperparameter search for our task-IG affinity model.\\
%\item 
$\bullet$ \textit{Extrinsic evaluation}: We have not yet performed extensive extrinsic evaluation on downstream tasks which use commonsense knowledge such as QA. We plan to do this as a future work.\\
%\item 
$\bullet$ \textit{Hardware constraints}: We are using smaller models (sentence BERT and NLI models) and also not fine-tuning the pre-trained models due to hardware constraints. %and for maintaining low carbon footprint. %Given enough hardware, we could have used larger pre-trained models and even explored fine-tuning of all or some of its layers.
%\end{enumerate}

\section{Related Work}
Most prior work formalizes the augmentation of task-IG knowledge as a classification problem. \citet{tagarev2019comparison} address the task of categorizing companies within industry classification schemes using encyclopedia articles. \citet{wood2017automated} present a deep neural-based industry classification to construct a database of companies labelled with the corresponding industry. \citet{davison2019commonsense} develop a technique to rank a triple by estimating pointwise mutual information between two entities using a large pre-trained bidirectional language model. \citet{distiawan2019neural} assumes the existence of the head and tail entities in KB and tries to find more relationships between them. Open Information Extraction (Open IE) paradigm~\cite{niklaus2018survey} is a related but a different line of research. Open IE is not limited to a pre-defined set of target relations, but extracts all types of relations found in a text. Our problem is different from Open IE in the sense that the triples that we want to extract, need not be explicitly mentioned in the text. In other words, even if the IG is not mentioned explicitly in the text, our goal is to infer the IG using the tasks which is not possible by Open IE techniques. We are trying to extract a highly semantic relation $\langle${\small\sf IG company}, is~capable~of, {\small\tt Task}$\rangle$, from sentences where IG may not be explicitly mentioned and where {\small\tt Task} is any voluntary business-specific action carried out by any organization in the IG. OpenIE does not extract any mention of any {\small\tt Task} (which is a complex semantic notion), nor can it extract any tuple for an IG when it is not explicitly mentioned in the sentence.

\section{Conclusions and Future Work}
%In this work, 
We advocated the augmentation of task-IG information in a commonsense knowledge base, by matching tasks with suitable IGs. We devised a framework to automatically extract task-IG information from natural language text that does not require any manually annotated training instances. %We release the top 100 tasks for 24 IGs to be added to ConceptNet.
We recommend 2339 triples to be added in ConceptNet whereas there exist only 138 triples of the form $\langle${\small\sf IG company}, is~capable~of, {\small\tt Task}$\rangle$, indicating addition of considerable new knowledge. %in ConceptNet.

The work opens a few strong research directions. Note that as the framework does not require any training instances, it can be extended to any number and type of text datasets. In future, we plan to address some of the limitations listed above. Especially, we plan to explore towards %extending 
generalizing the approach to other relation types in triples. Also, some other directions would be to look for role based information (e.g., $\langle$ {\small\sf Engineer}, {\it is~capable~of}, {\small\tt build bridges}$\rangle$) and to automatically estimate difficulty level of a task.


\bibliography{anthology,custom}
\bibliographystyle{acl_natbib}
%\appendix
%\section{Example Appendix}
%\label{sec:appendix}
%This is a section in the appendix.
\end{document}